\documentclass[letterpaper]{article}
\usepackage{aaai}
\usepackage{times}
\usepackage{helvet}
\usepackage{courier}
\usepackage{url}
\usepackage{relsize}
\usepackage{wrapfig}
\usepackage{balance}
\usepackage{array}
\usepackage{amsmath}
\usepackage{mathrsfs}
\usepackage{amssymb}
\usepackage{amsthm}
\usepackage{dsfont}
\usepackage[disable]{todonotes}
\newcommand{\todoj}[2][]{}
\newcommand{\todom}[2][]{}
\newcommand{\todoa}[2][]{}
\newcommand{\todoMGH}[2][]{}

\usepackage{algorithm}
\usepackage{algorithmic}

\usepackage{graphicx}

\newcommand{\R}{{\mathbb R}}

\newcommand{\transpose}{^\mathsf{\scriptscriptstyle T}}

\newcommand{\boldf}{\mathbf{f}}

\newcommand{\btheta}{{\boldsymbol \theta}}

\newcommand{\tSigma}{\widetilde{\Sigma}}

\newcommand{\beq}{\begin{equation}}
\newcommand{\eeq}{\end{equation}}

\newcommand{\beqa}{\begin{eqnarray}}
\newcommand{\eeqa}{\end{eqnarray}}

\newcommand{\beqan}{\begin{eqnarray*}}
\newcommand{\eeqan}{\end{eqnarray*}}

\newcommand{\beqal}{\begin{align*}}
\newcommand{\eeqal}{\end{align*}}

\begin{document}
%
\title{Maximum Entropy Semi-Supervised Inverse Reinforcement Learning}

\author{
 Julien Audiffren$^1$ \quad Michal Valko$^2$ \quad Alessandro Lazaric$^2$ \quad Mohammad Ghavamzadeh$^{2,3}$ \\
 $^1$CMLA, ENS Cachan \quad $^2$SequeL team, INRIA Lille - Nord Europe \quad $^3$Adobe Research \\
}

\maketitle


\begin{abstract}

A relatively recent approach to apprenticeship learning (AL) is to formulate it
as an inverse reinforcement learning (IRL) problem. 
MaxEnt-IRL successfully integrates the maximum entropy principle
into IRL, and unlike its predecessors, it resolves the
ambiguity arising from the fact that a possibly large number of policies could
match the expert's behavior. In this paper, we study an AL setting in which in
addition to the expert's trajectories,
a number of unsupervised trajectories is available, and we introduce a new
algorithm, called MESSI, that combines MaxEnt-IRL
with principles coming from semi-supervised learning. In particular, MESSI
integrates the unsupervised data into
the MaxEnt-IRL framework using a pairwise penalty on trajectories. Empirical
results in a highway driving and two grid-world  
problems indicate that MESSI is able to take advantage of the unsupervised trajectories and to improve the performance of
MaxEnt-IRL. 

\end{abstract}

\section{Introduction}

The most common approach to solve a sequential decision-making problem is to
formulate it as a Markov decision process (MDP) and then use a dynamic
programming or a reinforcement learning algorithm to compute a (near-)~optimal
policy. This process requires the definition of a reward function such
that the policy obtained by optimizing the resulting MDP produces the desired
behavior. However in many applications, such as driving or playing tennis, it is
required to take into account different desirable factors, and it might be
difficult to define an explicit reward function that accurately specifies the trade-off between these desiderata. It is often easier and more natural to learn how
to perform such tasks by observing an expert's demonstration. The task of
learning from an expert is called {\em apprenticeship learning} (also referred
to as {\em imitation learning}). A powerful and relatively novel approach to
apprenticeship learning (AL) is to formulate it as an {\em inverse 
reinforcement learning} (IRL) problem~\cite{ng2000algorithms}. The basic idea
is to assume that the expert is trying to optimize an MDP whose reward function
is unknown, and to derive an algorithm for learning the task demonstrated by the
expert~\cite{ng2000algorithms,abbeel2004apprenticeship}. This approach has been
shown to be effective in learning non-trivial tasks such as inverted helicopter
flight control~\cite{ng2004inverted},
ball-in-a-cup~\cite{boularias2011relative}, and driving on a
highway~\cite{abbeel2004apprenticeship,levine2011nonlinear}.

In the IRL approach to AL, we assume that several trajectories generated by an
expert are available and the {\em unknown} reward function optimized by the
expert can be specified as a linear combination of a number of state features. For each
trajectory, we define its {\em feature count} as the sum of the values of
each feature across the states traversed by the trajectory. The {\em expected feature count}, computed as the average of the feature counts of all the
expert trajectories, encodes the behavior of the expert, and thus,
the goal is to find policies,\footnote{More precisely, the goal is to find
reward functions such that the behavior of the expected feature
count of the (near)-optimal policies of the resulting MDPs match that of the
expert.} whose expected feature counts match that of the expert. The early
method proposed by Abbeel and Ng~\shortcite{abbeel2004apprenticeship} finds such policies (reward
functions) using an iterative max-margin algorithm. A major disadvantage of
this 
approach is that the problem is ill-defined since a possibly large number of
policies can satisfy this matching condition.
In this paper, we build on Ziebart et al.~\shortcite{ziebart2008maximum} that proposed
a method based on the maximum entropy (MaxEnt) principle to resolve this
ambiguity. 
A more comprehensive review of the IRL literature is available in
Sec.~\ref{sec:related} of the supplementary material.

In many applications, in addition to the expert's trajectories, we may have
access to a large number of trajectories that are not necessarily performed by an ``expert". For example, in learning to drive, we may ask an
expert driver to demonstrate a few trajectories and use them in an AL algorithm to
mimic her behavior. At the same time, we may
record trajectories from many other drivers for which we cannot assess their
quality (unless we ask an expert driver to evaluate them) and that may or may
not demonstrate an expert-level behavior.
We will refer to them as \textit{unsupervised trajectories}
and to the task of learning with them as \textit{semi-supervised apprenticeship
learning} following Valko et al.~\shortcite{valko2012semi-supervised}.
However, unlike in classification, we do not regard the
unsupervised trajectories as being a mixture of expert
and non-expert classes. This is because the unsupervised trajectories might have been
generated by the expert herself, by another expert(s), by near-expert agents, by agents
maximizing different reward functions, or simply they can be some noisy data.
The objective of IRL is to find the reward
function that expert trajectories maximize, and thus, semi-supervised apprenticeship learning cannot be considered as a special case of
semi-supervised classification.\footnote{If that was true, then the AL problem would
reduce to a classical supervised learning problem upon revealing the
expert/non-expert labels of the additional trajectories, which is not the case.} 
%
Similar to many IRL algorithms that draw inspiration from supervised learning,
in this paper, we build on semi-supervised learning
(SSL) (Chapelle et al.~\citeyear{chapelle2006semi-supervised}) tools to derive a novel version of
MaxEnt-IRL that takes advantage of unsupervised trajectories to improve the
performance. In Valko et al.~\shortcite{valko2012semi-supervised}, the authors proposed a semi-supervised
apprenticeship learning paradigm, called SSIRL, that combines the IRL approach
of Abbeel and Ng~\shortcite{abbeel2004apprenticeship} with semi-supervised SVMs,
and showed how it could take advantage of the unsupervised data
and perform better and more efficiently (with a smaller number of iterations, and
thus, solving fewer MDPs) than the original algorithm
of Abbeel and Ng~\shortcite{abbeel2004apprenticeship}. Nonetheless, as we will discuss in Sec.~\ref{sec:ssal}, SSIRL still suffers from the ill-defined nature
of this class of IRL methods and the fact that all the policies (reward functions)
generated over the iterations are labeled as ``bad". This creates more
problems in the SSL case than in the original setting. In fact, SSIRL relies on a
cluster assumption that assumes the good and bad trajectories are
well-separated in some feature space, and that never holds as new policies are generated and labeled as ``bad''. Finally, even the authors admit that SSIRL
is rather a proof of concept than an actual algorithm, as it does
not have a stopping criterion, and over iterations the performance always reverts
back to the performance of the basic algorithm
by Abbeel and Ng~\shortcite{abbeel2004apprenticeship}.
%
In Sec.~\ref{sec:pairwise}, we address these problems
by combining the MaxEnt-IRL approach of Ziebart et al.~\shortcite{ziebart2008maximum} with SSL. We
show how the unsupervised trajectories can be integrated into the MaxEnt-IRL
framework in a principled way such that the resulting algorithm, MESSI
(MaxEnt Semi-Supervised IRL), performs better than
MaxEnt-IRL. Finally, in Sec.~\ref{s:experiments}, we empirically show this improvement in 
the highway driving problem of Syed et al.~\shortcite{syed2008apprenticeship}. 
\section{Background and Notations}\label{s:background}
\label{sec:bck}


A Markov decision process (MDP) is a tuple $\langle S, A, r,
p\rangle$, where $S$ is the state space, $A$ is the action space, $r:
S\rightarrow \R$ is the state-reward function, and $p: S\times A \rightarrow
\Delta(S)$ is the dynamics, such that $p(s'|s,a)$ is the probability of reaching
state $s'$ by taking action $a$ in state $s$. We also denote by $\pi:
S\rightarrow \Delta(A)$ a (stochastic) policy, mapping states to distributions
over actions. As customary in IRL, we assume that the reward function can be
expressed as a linear combination of state-features. Formally, let $\boldf: S
\rightarrow \R^d$ be a mapping from the state space to a $d$-dimensional feature
space and let $\boldf^i$ denote the $i$-th component of the vector $\boldf$,
then we assume that the reward function is fully characterized by a vector
$\btheta \in \R^d$ such that for any $s\in S$, $r_\btheta(s) = \btheta\transpose
\boldf(s)$. A trajectory $\zeta = (s_1,a_1,s_2,\ldots, a_{T-1},s_T)$ is a
sequence of states 
and actions,\footnote{In practice, MaxEnt-IRL and our extension do not require a trajectory to include actions, only states.} whose cumulative reward is

\vspace{-0.1in}
\begin{small}
\begin{align}\label{eq:gen.model}
\bar r_\theta(\zeta) = \sum_{t=1}^T r_\theta(s_t) = \sum_{t=1}^T \btheta\transpose \boldf(s_t) = \btheta\transpose \sum_{t=1}^T\boldf(s_t) = \btheta\transpose \boldf_{\zeta},
\end{align}
\end{small}
\vspace{-0.05in}

\noindent where $\boldf_{\zeta}$ is the \textit{feature count} of trajectory $\zeta$ that measures the cumulative ``relevance'' of each of the components $\boldf^i$ of the feature vector  along the states traversed by $\zeta$.\footnote{Note that if $\boldf : S \rightarrow \{0,1\}$, then $\boldf_{\zeta}^i$ reduces to a counter of the times $\zeta$ traverses states activating the $i$th feature.} 

In AL, we assume that only $l$ trajectories $\Sigma^* =
\{\zeta_i^*\}_{i=1}^l$ are demonstrated by the expert and we define the
corresponding \textit{expected} feature count as $\boldf^* = \frac{1}{l}
\sum_{i=1}^l \boldf_{\zeta_i^*}$. The expected feature count $\boldf^*$
effectively summarizes the behavior of the expert since it defines which
features are often ``visited'' by the expert, and thus, implicitly revealing her
preferences, even without an explicit definition of the reward. This evidence is
at the basis of the expected feature-count matching idea (see
e.g.,~Abbeel and Ng~\citeyear{abbeel2004apprenticeship}), which defines the objective of
AL as finding a policy $\pi$ whose corresponding
distribution $P_{\pi}$ over trajectories is such that  the expected feature
count of $\pi$ matches the expert's, i.e., 
%
\begin{align}\label{eq:feat.matching}
\sum_{\zeta} P_{\pi}(\zeta)\boldf_{\zeta} = \boldf^*,
\end{align}
%
where the summation is taken over all possible trajectories. As pointed out
by Ziebart et al.~\shortcite{ziebart2008maximum}, there is a large number of policies that
satisfy the matching condition in Eq.~\ref{eq:feat.matching}. Thus, they
introduce a method to resolve this ambiguity and to discriminate between
different policies $\pi$ based on the actual reward they accumulate, so that the
higher the reward the higher the probability. Building on
the maximum entropy principle, Ziebart et al.~\shortcite{ziebart2008maximum} move from the space of
policies directly to the space of trajectories and define the probability of a
trajectory $\zeta$ in a (stochastic) MDP characterized by a reward function
defined by $\btheta$ as
\begin{align*}
P(\zeta|\btheta) \approx \frac{\exp(\btheta\transpose f_\zeta)}{Z(\btheta)} \prod_{t=1}^T p(s_{t+1}|s_t, a_t),
\end{align*}
where $Z(\btheta)$ is a normalization constant. The policy $\pi$ corresponding to the distribution induced by the reward $\btheta$ is
\begin{align}\label{eq:policy}
\pi(a| s; \btheta) = \sum_{\zeta\in \Sigma_{s,a}} P(\zeta|\btheta),
\end{align}
where $\Sigma_{s,a}$ denotes the set of trajectories for which action $a$ is taken in state $s$. Given this formulation of the probability of the trajectories, the MaxEnt-IRL algorithm searches for the reward vector $\btheta$ that maximizes the likelihood of the expert trajectories. In particular, MaxEnt-IRL maximizes the log-likelihood of $\btheta$ given the set of expert trajectories $\Sigma^* = \{\zeta_i^*\}_{i=1}^l$, i.e.,
\begin{align}\label{eq:maxent.opt}
\btheta^* = \arg\max_{\btheta} L(\btheta|\Sigma^*) = \arg\max_{\btheta} \sum_{i=1}^l \log P(\zeta_i^*|\btheta).
\end{align}
The resulting reward function is such that the corresponding distribution over trajectories, $\sum_{\zeta} P(\zeta|\btheta^*)\boldf_{\zeta}$, matches the expected feature count $\boldf^*$, and at the same time, it strongly penalizes trajectories (and implicitly policies) that do not achieve as much reward as the expert.


\section{Maximum Entropy Semi-Supervised Inverse Reinforcement Learning}
\label{sec:ssal}


The main objective of \textit{semi-supervised learning}
(SSL) (Chapelle et al.~\citeyear{chapelle2006semi-supervised}) is to bias the learning toward models
that assign similar outputs to intrinsically similar data. Similarity is commonly formalized in SSL as \textit{manifold}, \textit{cluster},
or other \textit{smoothness} assumptions. In Valko et al.~\shortcite{valko2012semi-supervised}, a
semi-supervised regularizer is integrated into the SVM-like structure of the
feature matching algorithm of Abbeel and Ng~\shortcite{abbeel2004apprenticeship} under an implicit
clustering assumption. Although the integration is technically sound, the SSL
hypothesis of the existence of clusters grouping trajectories with different
performance is not robust w.r.t.~the learning algorithm. In fact, even
assuming that expert and non-expert policies (used to generate
part of the unsupervised trajectories) are actually clustered, newly generated
policies are always labeled as ``negative'', and as they become more and more
similar to the expert's
policy, they will eventually cross the clusters and invalidate the SSL
hypothesis.

In this section, we propose a principled SSL solution to AL
with unsupervised trajectories. To avoid the problems
of the prior work, we do not reduce the problem to SSL classification.
We build on the SSL work of Erkan and Altun~\shortcite{erkan2009semi-supervised} that shows how unsupervised data can be
integrated into the maximum entropy framework via modifications of the
optimization problem that reflect the structural assumptions of the geometry of the data. In
our context, a similar approach is applied to MaxEnt-IRL that leads to
constraining the probabilities of the trajectories to be
\textit{locally consistent}.

\subsection{Pairwise Penalty and Similarity Function}
\label{sec:pairwise}


We assume that the learner is provided with a set of expert trajectories $\Sigma^*=\{\zeta_i^*\}_{i=1}^l$ and a set of unsupervised trajectories $\tSigma=\{\zeta_j\}_{j=1}^u$. We also assume that a function $s$ is provided to measure the similarity $s(\zeta,\zeta')$ between any pair of trajectories $(\zeta,\zeta')$. We define the \textit{pairwise penalty $R$} as
\begin{align}\label{eq:penalty}
 R(\btheta|\Sigma) = \frac{1}{2(l+u)}\sum_{\zeta, \zeta'\in \Sigma}
s\left(\zeta,\zeta'\right)
(\btheta\transpose(\boldf_{\zeta} - \boldf_{\zeta'}))^2,
\end{align}
where $\Sigma = \Sigma^*\cup\tSigma$, and $\boldf_{\zeta}$ and $\boldf_{\zeta'}$ are the feature counts for trajectories $\zeta,\zeta'\in\Sigma$, and finally $(\btheta\transpose (\boldf_{\zeta} - \boldf_{\zeta'}))^2 = (\bar r_\btheta(\zeta) - \bar r_\btheta(\zeta'))^2$ is the difference in rewards accumulated by the two trajectories w.r.t.~the reward vector $\btheta$.
The purpose of the pairwise penalty is to penalize reward vectors $\btheta$ that
assign very different
rewards to similar trajectories (as measured by $s(\zeta,\zeta')$). In other words, the pairwise penalty acts as a regularizer which favors
vectors that encode giving similar rewards to similar trajectories. Note
also that whenever two trajectories are very different, $s(\zeta,\zeta')$ is
very small, and this allows us to choose any $\btheta$ that gives very different
rewards.

In SSL, there are several ways to integrate a smoothness penalty $R(\btheta)$
into the learning objective. However, if the objective is that of
maximum entropy, such as in MaxEnt-IRL, the method of Erkan and Altun~\shortcite{erkan2009semi-supervised}
offers a principled way for regularization. In our specific case, this
corresponds to adding the pairwise penalty $R(\btheta)$ to the MaxEnt-IRL
optimization problem of Eq.~\ref{eq:maxent.opt} as
\begin{align}\label{eq:ssl.opt}
\btheta^* = \arg\!\max_\btheta \ \left(L(\btheta | \Sigma^*)  -  \lambda R(\btheta|\Sigma)\right),
\end{align}
where $\lambda$ is a parameter trading off between the log-likelihood of
$\btheta$ w.r.t.~the expert's trajectories and the coherence with the similarity
between the provided trajectories both in $\tSigma$ and $\Sigma^*$.
As it is often the case in SSL, the choice of the
similarity is critical to the success of learning. It encodes a prior on the
characteristics two trajectories should share when
they have similar performance, and it defines how unsupervised data together with
the expert data are used in
regularizing the final reward. Whenever enough prior knowledge about the
problem is available, it is possible to hand-craft specific similarity functions
that effectively capture the critical characteristics of the trajectories
(see the experiments in Sec.~\ref{sub:pit} of the supplementary material). On the other hand, when the knowledge about the problem
is rather limited, it is preferable to use generic similarity functions that
can be employed in a wide range of domains. A popular choice of a generic
similarity is the radial basis function (RBF) kernel
$s\left(\zeta,\zeta'\right) = \exp(-\| \boldf_{\zeta} - \boldf_{\zeta'}\|^2 / 2 \sigma )$,
where $\sigma$ is the bandwidth. Although hand-crafted similarity
functions usually perform better, we show in Sec.~\ref{s:experiments} that
even the simple RBF is an effective similarity for the feature counts.

More generally, the effectiveness of the regularizer
$R(\btheta|\Sigma)$ is not only related to the quality of the similarity
function but also to the unsupervised trajectories. To justify this claim,
let $P_u$ be a probability distribution over the set $\overline{\Sigma}$ of all
the feasible trajectories (i.e., trajectories compatible with the MDP
dynamics); we assume that the unsupervised trajectories are drawn from $P_u$.
Note that the generative model of the unsupervised trajectories could be
interpreted as the result of a mixture of behaviors obtained from different
reward functions. Let $P(\theta)$ be a distribution over reward functions (e.g.,
a set of users with different preferences or skills), then
$P_u$ is simply obtained as $P_u(\zeta) = P(\zeta|\theta)P(\theta)$. Whenever a
very informative similarity function is available, then the impact of $P_u$ on
the performance is rather limited (see the pit problem in
Sec.~\ref{sub:pit} of the supplementary material).\footnote{Note that if  $s$ is not smooth w.r.t.\,the feature counts, MESSI would perform
 worse than MaxEnt-IRL. In general, when limited knowledge
of the trajectories' similarity is available, it is preferable
to define $s$ with a rather small generalization across
trajectories.} On the other hand, when $s$ only provides local similarity (e.g.,~RBF), it is the distribution $P_u$
that defines the way the similarity is propagated across trajectories, since $s$ is only applied to trajectories in $\Sigma$ drawn from $P_u$.

We now give some intuition of the effect of the unsupervised trajectories
through the similarity function.
Let us consider an (uninteresting) case when $P_u$ is uniform over
$\overline{\Sigma}$ and RBFs are used with a small bandwidth $\sigma$.
This would reduce to a regularized version of
MaxEnt-IRL with a $L_2$-regularization on the parameter $\btheta$ that is
forced to be \textit{uniformly} smooth over all trajectories in
$\overline{\Sigma}$.
However, in the typical case, we expect $P_u$ to be
non-uniform, which enforces the regularization towards $\btheta$'s that give
similar rewards only to similar trajectories among those that are more likely to
be present. Similar to the regularization proposed in the SSL
method of Erkan and Altun~\shortcite{erkan2009semi-supervised}, here we expect that the more
$P(\cdot|\btheta^*)$ (i.e., the trajectory distribution induced by the reward maximized by
the expert) is supported\footnote{Here we use \textit{support} as in the measure
theory. Intuitively, it means that the trajectories that are likely
to be sampled from $P(\cdot|\btheta^*)$ are present in $P_u$.} in $P_u$, the
more effective the penalty $R(\btheta|\Sigma)$. Since the regularization depends
on
$\btheta$, in this case, MESSI forces the similarity only among the trajectories
that are likely to perform well.
Finally, similar to standard SSL, if $P_u$ is rather adversarial, i.e.,~if
the support of $P(\zeta|\theta^*)$ is not supported in $P_u$, then penalizing
according to the unsupervised trajectories may even worsen the performance. In
the experiments reported in the next section, we will define $P_u$ as a mixture
of $P(\cdot|\theta^*)$ (used to generated the expert trajectories) and
trajectories generated by other distributions, and show that the unfavorable case has only a limited effect on the MESSI's performance.

\vspace{-0.1in}
\subsection{Implementation of the Algorithm}
\label{sec:opti}
\vspace{-0.2em}

\begin{algorithm}[t]
\begin{small}
\caption{MESSI - MaxEnt SSIRL}
 \label{alg:SSL.MaxEnt}
 \begin{algorithmic}
 \STATE {\bfseries Input:} Set of $l$ expert trajectories $\Sigma^*=\{\zeta^*_i\}_{i=1}^l$, set of $u$ unsupervised trajectories $\tSigma =\{\zeta_j\}_{j=1}^u$, similarity function $s$, number of iterations $T$, constraint $\theta_{\max}$, regularizer $\lambda_0$
     \STATE {\bfseries Initialization:}
 \STATE Compute $\{\boldf_{\zeta^*_i}\}_{i=1}^l$, $\{\boldf_{\zeta_j}\}_{j=1}^u$ and $\boldf^* = 1/l \sum_{i=1}^l \boldf_{\zeta^*_i}$
 \STATE Generate a random reward vector $\btheta_0$
     \FOR {$t = 1$ {\bfseries to}  $T$} 
   \STATE  1. Compute policy $\pi_{t-1}$ from $\btheta_{t-1}$ (backward pass Eq.~\ref{eq:policy})
   \STATE  2. Compute feature counts $\boldf_{t-1}$ of $\pi_{t-1}$ (forward pass Eq.~\ref{eq:feat.count.t})
   \STATE  3. Update the reward vector as in Eq.~\ref{eq:update}
    \STATE 4. If $\| \btheta_t \|_{\infty} > \theta_{\max}$, project back by $\btheta_t  \gets \btheta_t \frac{\theta_{\max}}{\|\btheta_t\|_\infty}$
    \ENDFOR
 \end{algorithmic}
 \end{small}
\end{algorithm}

Alg.~\ref{alg:SSL.MaxEnt} shows the pseudo-code of MESSI:
\textit{M}ax\textit{E}nt \textit{S}emi-\textit{S}upervised \textit{I}RL.
MESSI solves the optimization problem of Eq.~\ref{eq:ssl.opt} by gradient
descent. At the beginning, we first compute the empirical feature counts of all
the expert's $\{\boldf_{\zeta^*_i}\}$ and unsupervised trajectories $\{\boldf_{\zeta_j}\}$, and randomly initialize the reward vector $
\btheta_0$. At each iteration $t$, given the reward $\btheta_t$, we
compute the corresponding expected feature count $\boldf_t$ and obtain
$\btheta_{t+1}$ by the following update rule:
\begin{equation}\label{eq:update}
\begin{aligned}
\btheta_{t+1} &= \btheta_{t} +  (\boldf^* - \boldf_t) \\
&+ \frac{\lambda}{\theta_{\max}(l+u)} \sum_{\zeta,\zeta'\in\Sigma} s(\zeta,\zeta')
\left(\btheta_{t}\transpose
(\boldf_{\zeta}-\boldf_{\zeta'})\right)^2,
 \vspace{-0.1em}
\end{aligned}
\end{equation}
where $\boldf^*$ is the average feature count of the expert trajectories in $\Sigma^*$ and $\theta_{\max}$ is a parameter discussed later. A critical step in MESSI is the computation of the expected feature count $\boldf_t$ corresponding to the given reward vector $\btheta_t$. In fact, $\boldf_t$ is defined as $\sum_{\zeta} P(\zeta|\btheta_t) \boldf_\zeta$ and it requires the computation of the ``posterior'' distribution $P(\zeta|\btheta_t)$ over all the possible trajectories. This becomes rapidly infeasible since the number of possible trajectories grows exponentially with the number of states and actions in the MDP. Thus, we follow the same approach illustrated in Ziebart et al.~\shortcite{ziebart2008maximum}. We note that the expected feature count can be written as
\begin{align}\label{eq:feat.count.t}
\boldf_t = \sum_{\zeta} P(\zeta|\btheta_t) \boldf_\zeta = \sum_{s\in
S}\rho_t(s)\boldf(s),
 \vspace{-0.2em}
\end{align}
where $\rho_t(\cdot)$ is the expected visitation frequency of the states $s\in S$ obtained from following the policy induced by $\btheta$. As a result we first need to compute the policy $\pi_t$ as in Eq.~\ref{eq:policy}. This can be done using a value-iteration-like algorithm as in the \textit{backward pass} of Alg.~1 in Ziebart et al.~\shortcite{ziebart2008maximum}.\footnote{This step is the main computational bottleneck of MESSI. It is directly inherited from MaxEnt-IRL and is common to most IRL methods. However, using the transition samples in $\Sigma$, we may apply an approximate dynamic programming algorithm, like fitted value iteration, to reduce the computational complexity and remove the need for knowing the exact dynamics of the MDP.} Once the stochastic policy $\pi_t$ is computed, starting from a given initial distribution over $S$, we recursively apply $\pi_t$ and using the transition model $p$ compute the expected visitation frequency $\rho$ (\textit{forward pass}).
%
%
Although the backward pass step of MaxEnt-IRL allows the estimation of $\pi_t$,
it requires computing the exponential of the reward associated to the feature
count of any given path. This may introduce numerical issues that may prevent
the algorithm from behaving well. Thus, we introduce two modifications to the
structure of the MaxEnt-IRL algorithm. Since these
issues exist in the original MaxEnt-IRL and are not caused by the SSL
penalty, the following two modifications could be integrated into the original
MaxEnt-IRL as well.

We first normalize all the features so that for any $s$,
$\boldf(s)\in [0,1]^d$. We then multiply them by $(1-\gamma)$ and move to the discounted feature
count. In other words, given a trajectory $\zeta=(s_1,\ldots,s_T)$, if the
features $\boldf$ are multiplied by $(1-\gamma)$, we have that the discounted
feature count of $\zeta$, $\boldf_{\zeta} = \sum_{t=0}^{\infty} \gamma^t
\boldf(s_t)$, is bounded in $[0,1]$, i.e., $ \| \boldf \|_\infty \leq 1$. As a
result, all the expected feature counts involved in the optimization problem will
be bounded.
Although normalizing features already provides 
more stability to the backward pass, the cumulative rewards $\bar
r_{\btheta}(\zeta)$ still depend on $\btheta$, which would often tend to be
unbounded. In fact, if the expert trajectories often visit feature $\boldf^i$
while completely avoiding feature $\boldf^j$, then we want apprenticeship
learning to find a reward vector $\btheta$ that favors policies that reach
$\boldf^i$ much more often than $\boldf^j$. However, from
Eq.~\ref{eq:policy} we note that the probability of an action $a$ in a state
$s$ is obtained by evaluating the \textit{exponential} rewards accumulated by
different
trajectories that take action $a$ in state $s$. So, in order to obtain a policy
$\pi(\cdot|\cdot;\theta)$ that has a strong tendency of reaching $\boldf^i$ and
avoiding $\boldf^j$, the reward vector $\btheta$ should be very positive for
$\btheta^i$ and very negative for $\btheta^j$. This may lead to very extreme
reward vectors such that $\btheta_t^i \rightarrow \infty$ and $\btheta_t^j
\rightarrow -\infty$ as the algorithm proceeds, which would rapidly lead to numerical instability. In order to prevent this effect, we introduce a constraint $\theta_{\max}$ such that $||\btheta_t||_\infty \leq \theta_{\max}$. From an algorithmic point of view, the introduction of the constraint $\theta_{\max}$ requires ensuring that at the end of each iteration of the gradient descent, the $\btheta$ obtained after the update is always projected back into the set of $||\btheta||_\infty \leq \theta_{\max}$. Finally, it is sensible to use the constraint to tune the range of the regularizer $\lambda$ in advance, which should be set to $\lambda = \lambda_0 / \theta_{\max}$. The constraint $\theta_{\max}$ can also be seen as the minimum entropy (or uncertainty) we would like to observe in the resulting policy. In fact, a very small $\theta_{\max}$ forces $\pi(\cdot|s;\theta)$ to have a very large entropy, where most of the actions have similar probability. Therefore, the parameter $\theta_{
\max}$ could 
be 
used by the designer to encode how much expert trajectories, that are realizations of the expert's policy, should be trusted, and how much residual
\textit{uncertainty} should be preserved, in the final policy $\pi(\cdot|s;\theta_T)$.

\vspace{-0.5em}
\section{Experimental Results}\label{s:experiments}
 
Here we report the empirical results of MESSI
and investigate to what extent
unsupervised trajectories are exploited by MESSI to learn better policies. We first describe the general structure of the experiments, including the grid-world problems reported in Sec.~\ref{s:add.experiments} of the supplementary material.

In order to have a sensible comparison between MESSI and
MaxEnt-IRL, we first need to define how unsupervised trajectories are actually
generated (i.e.,~the distribution $P_u$). We consider three cases, where the unsupervised
trajectories obtained from near-expert policies, random policies, and policies
optimizing different reward functions. Let $\btheta^*$ denote the reward
associated with the expert and $\btheta_1$ and $\btheta_2$ denote two other
rewards. We define  $P_{u^*} = P(\cdot \vert \btheta^*)$, $P_{1} = P(\cdot
\vert \btheta_1)$, and $P_{2} = P(\cdot \vert \btheta_2)$, and draw unsupervised trajectories from $P_{\mu_1}= \nu P_{u^*} +
(1-\nu)P_{1}$, $P_{\mu_2}= \nu P_{u^*} + (1-\nu)P_{2}$, and $P_{\mu_3}= \nu
P_{1} + (1-\nu)P_{2}$, where $\nu \in \left[0,1 \right]$ is a parameter. We
also consider MESSIMAX, the MESSI algorithm in which the unsupervised
trajectories are obtained from $P_{u^*}$. This is a favorable scenario used as
an upper-bound to the best attainable
performance of MESSI.
Because of the conceptual and practical issues discussed in Sec.~\ref{sec:ssal}, we do not compare with SSIRL, but rather compare to a semi-supervised baseline inspired by the EM algorithm (see e.g.,~Sec.~2 in Zhu~\citeyear{zhu2005semi-supervised}). The EM-MaxEnt starts from an arbitrary reward $\theta^0$ and at each round $K$ performs two steps:
\begin{enumerate}
\item \textit{Expectation step}: given $\btheta^{K-1}$, we compute the probabilities $P(\zeta|\btheta^{K-1})$ for each $\zeta\in\Sigma$.
\item \textit{Maximization step}: we solve a modified version of Eq.~\ref{eq:maxent.opt} where trajectories are weighted by their probabilities, i.e.,
\begin{align}\label{eq:em.maxent.opt}
\btheta^K = \arg\max_{\btheta} \sum_{\zeta\in\Sigma} P(\zeta|\btheta^{K-1}) \log P(\zeta_i^*|\btheta).
\end{align}
\end{enumerate}
In practice, at each round, Eq.~\ref{eq:em.maxent.opt} is solved using a gradient descent as in Alg.~\ref{alg:SSL.MaxEnt}. Therefore, we introduce an additional parameter $\eta$ which determines the number of gradient steps per-round in the maximization step. The resulting algorithm is referred to as $\eta$-EM-MaxEnt.

\begin{figure}
     \begin{center}
 \includegraphics[trim = 3mm 1mm 20mm 1mm,scale=0.22]{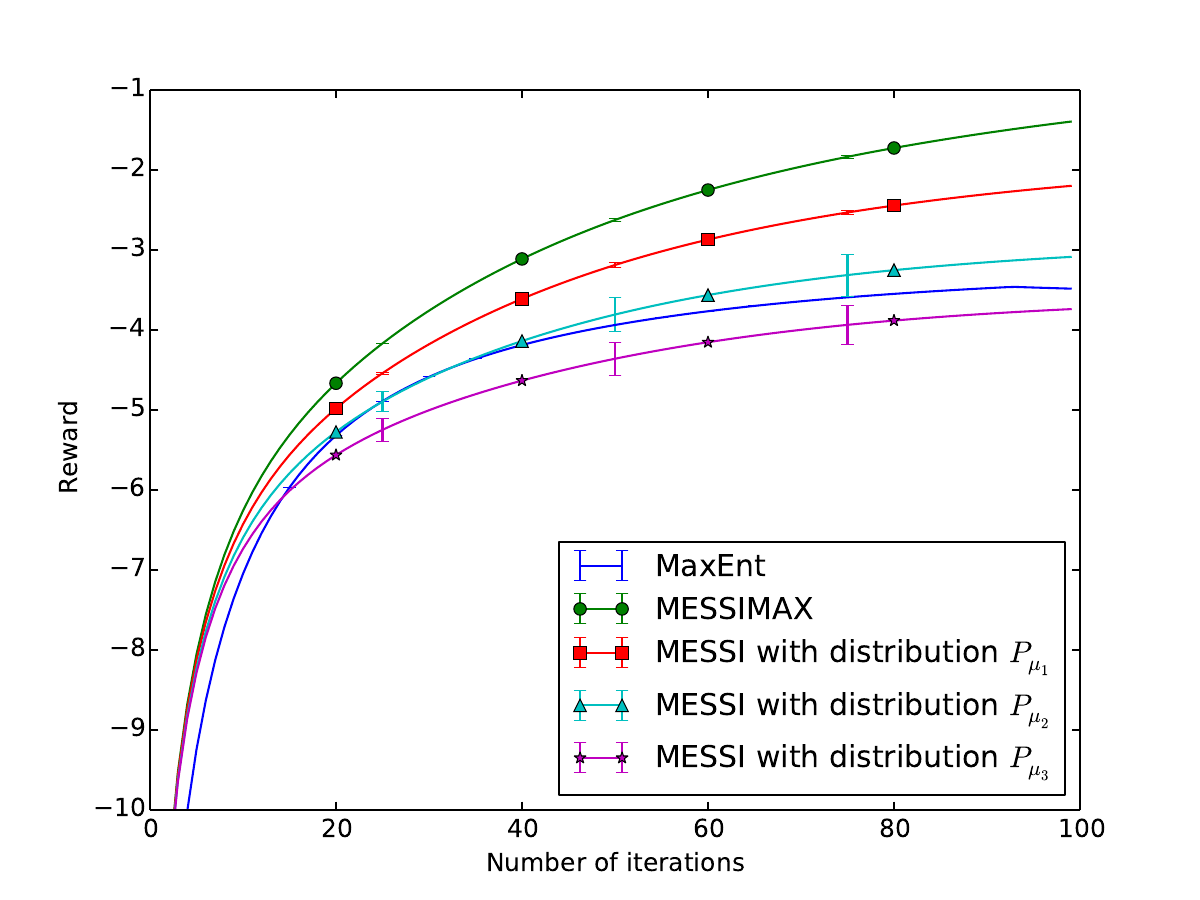}
 \includegraphics[trim = 1mm 1mm 5mm 1mm,scale=0.22]{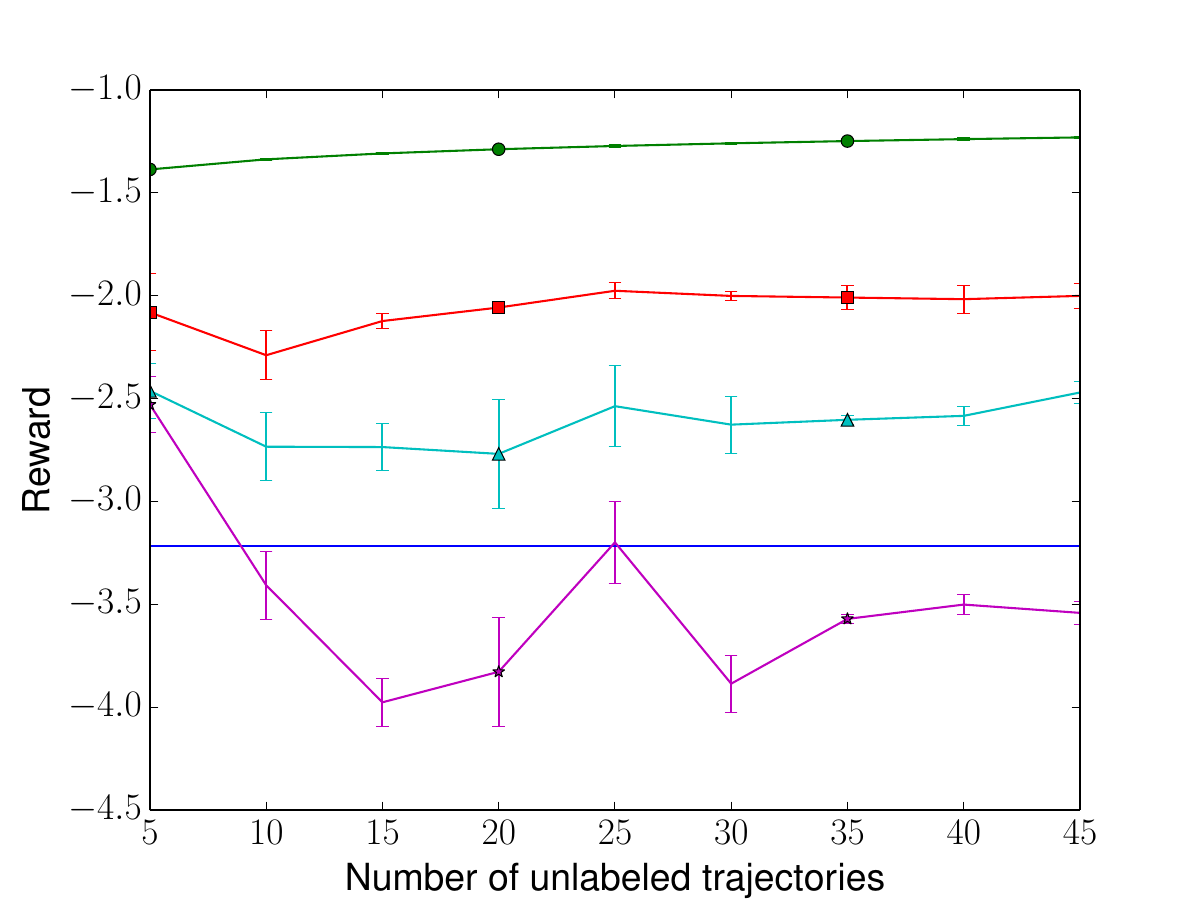}
 \includegraphics[trim = 15mm 1mm 10mm 1mm,scale=0.22]{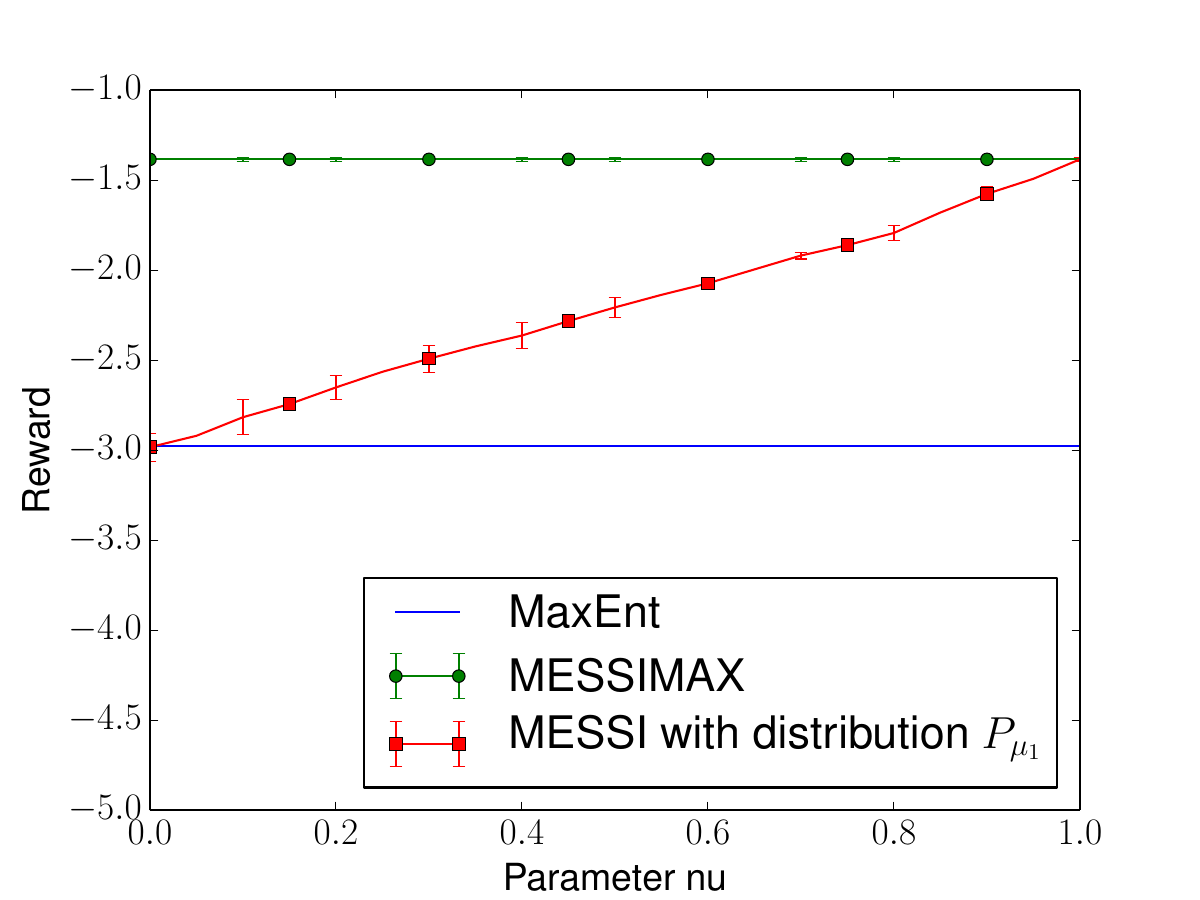}
 \includegraphics[trim = 1mm 1mm 30mm 1mm,scale=0.22]{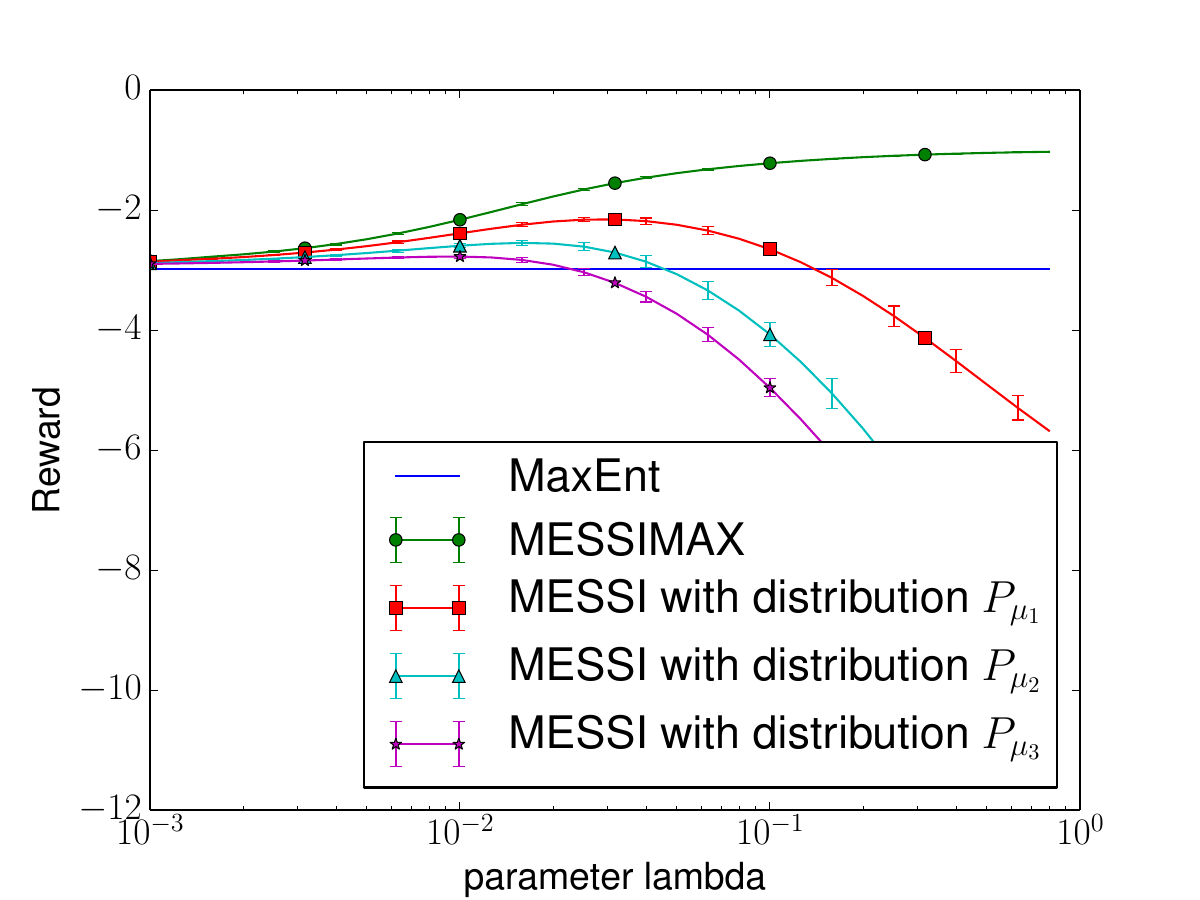}
 \caption{Results as a function of (from left to right): number of iterations of the algorithms, number of unsupervised trajectories, parameter $\nu$, and parameter $\lambda$.}
\label{fig:xphigh}
\vspace{-0.25in}
  \end{center}
 \end{figure}

\begin{figure}[t]
     \begin{center}
 \includegraphics[trim = 25mm 1mm 10mm 1mm,scale=0.1]{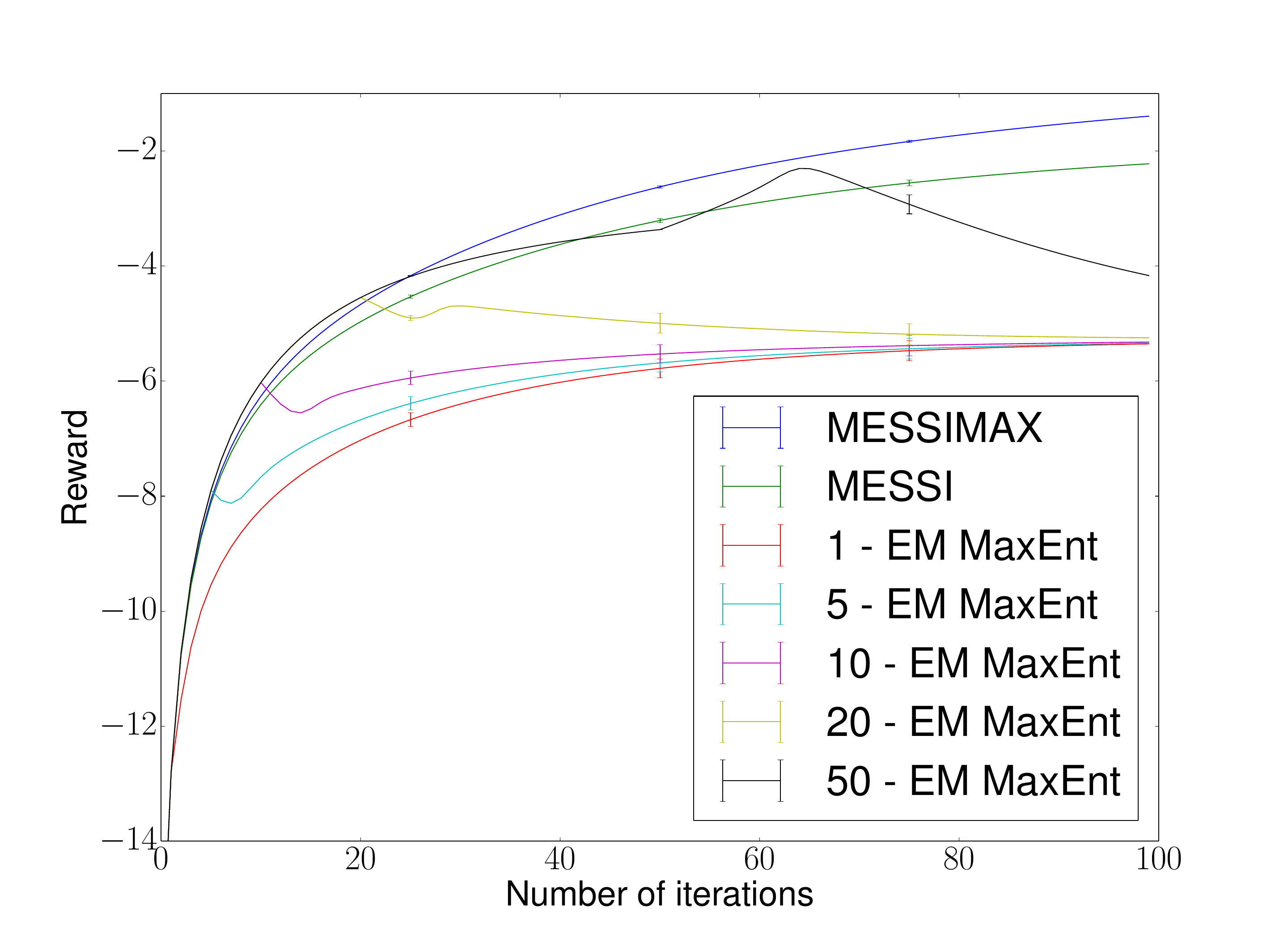}
\caption{Comparison of MESSI with MESSIMAX and $\eta$-EM-MaxEnt with $\eta=$1, 5, 10, 20, 50.}
\label{fig:comparison.em.drive}
  \end{center}
 \end{figure}

\textbf{Parameters.}
For each of the experiments, the default parameters are
$\theta_{\max}=500$, $\lambda_0=0.05$, the number of iterations of gradient
descent is set to $T=100$, one expert trajectory is provided $(l=1)$, and the
number
of unsupervised trajectories is set to $u=20$ with $\nu=0.5$.
In order to study the impact of different parameters on the algorithms, we first report results where we fix all the above parameters except one and
analyze how the performance changes for different values of the selected
parameter. In particular, in Fig.~\ref{fig:xphigh}, while keeping all the other parameters constant, we
compare the performance of MaxEnt-IRL, different setups of MESSI, and MESSIMAX
by varying different dimensions: \textbf{1)} the number of iterations of the
algorithms, \textbf{2)} the number of unsupervised trajectories, \textbf{3)} the
distribution $P_{1}$ by varying $\nu$, and \textbf{4)} the value of parameter $\lambda$. We then report a comparison with the EM-MaxEnt algorithm in Fig.~\ref{fig:comparison.em.drive}.

\textbf{The highway problem.}
We study a variation of the car driving problem in Syed~\shortcite{syed2008apprenticeship}.
The task is to navigate a car in a busy $4$-lane highway. The $3$ available actions are to move \textit{left}, \textit{right}, and \textit{stay} on the same lane. We use $4$ features: \textit{collision} (contact with another car), \textit{off-road} (being outside of the $4$-lane), \textit{left} (being on the two left-most lanes of the road), and \textit{right} (being on the two right-most lanes of the road). 
The expert trajectory avoids any collision or driving off-road while staying on
the left-hand side. The true reward $\btheta^*$ heavily penalizes
any collision or driving off-road but it does not
constrain the expert to stay on the left-hand side of the road. $\btheta_1$
penalizes any collision or driving off-road but with weaker weight, and finally
$\btheta_2$ does not penalize collisions.
We use the RBF kernel with $\sigma=5$ as the similarity function and evaluate the performance of a policy as \textit{minus} the expected number of collisions and off-roads.   





\textbf{Results.}
In the following, we analyze the results obtained by averaging 50 runs along multiple dimensions. The outcomes of the analysis are consistent with those in Sec.~\ref{s:add.experiments} of the supplementary material for the two grid-world problems.
 
\textit{Number of iterations.} As described in Sec.~\ref{sec:bck}, MaxEnt-IRL
solves the ambiguity of expected feature count matching by encouraging reward
vectors that strongly discriminate from expert's behavior and other policies.
However, while in MaxEnt-IRL, all trajectories that differ from the expert's are
(somehow) equally penalized, MESSI aims to provide similar rewards to
the (unsupervised) trajectories that resemble (according to the similarity function) the expert's. As a result, MESSI is able to find rewards that
better encode the unknown expert's preferences when provided with relevant unsupervised data. 
This intuitively explains the improvement of both MESSI and
MESSIMAX w.r.t.~MaxEnt-IRL. In particular, the benefit of
MESSIMAX, which only uses trajectories drawn from a distribution coherent with the expert's distribution, becomes apparent after just
a few dozens of iterations. 
On the other hand, the version of MESSI using trajectories drawn from
$P_{\mu_3}$, which is completely different from $P_{u^*}$, performs worse than
MaxEnt-IRL. As discussed in Sec.~\ref{sec:pairwise}, in this case, the
regularizer resulting from the SSL approach is likely to bias the
optimization away from the correct reward. Likewise, the versions of MESSI
using trajectories drawn from $P_{\mu_1}$ and $P_{\mu_2}$ lose part of the
benefit of having a bias toward the correct reward, but still perform better
than MaxEnt-IRL.
Another interesting element is the fact
that for all MaxEnt-based algorithms
the reward $\btheta_t$ tends to increase with the number of iterations (as described in Sec.~\ref{sec:opti}).
As a result, after a certain number of iterations, the reward reaches
the constraint $ \| \btheta \|_\infty = \theta_{\max}$, and when this happens,
additional iterations no
longer bring a significant improvement. In the case of MaxEnt-IRL, this may also cause overfitting and degrading performance.

\textit{Number of unsupervised trajectories.} The performance tends to improve as
more unsupervised trajectories are provided. This is natural
as the algorithm obtains more information about the features traversed by (unsupervised) trajectories that resemble the expert's. This allows MESSI to better generalize the expert's behavior over the entire state space. However in some experiments, the improvement
saturates after a certain number of unsupervised trajectories are added.
This effect is completely reversed when $P_{\mu_3}$ is used. In fact, in this case, the unsupervised trajectories do not provide any information about the behavior of the expert and the more the trajectories, the worse the performance.

\textit{Proportion of good unsupervised trajectories.} As in the case of
$P_{\mu_3}$, whenever provided with a completely non-relevant distribution,
MESSI may  perform
worse than MaxEnt-IRL. This is expected since the unsupervised
trajectories together with the similarity function and the pairwise penalty
introduce an inductive bias that forces similar trajectories to have similar
rewards. 
However in our problems, the results
show that this decrease of the performance is not dramatic and is indeed often negligible as soon as the distribution of the unsupervised trajectories supports the expert distribution $P_{u^*}$ enough. 
In fact, MESSI provided with trajectories from $ P_{\mu_1}$ and $ P_{\mu_2}$ starts performing better than MaxEnt-IRL when at least $\nu \ge 0.15$. 
Finally, note that when $\nu=1$, MESSI
and MESSIMAX are the same algorithms.
 
\textit{Regularization $\lambda$.} The improvement of MESSI over
MaxEnt-IRL depends on a proper choice of $\lambda$.
This is a well-known trade-off in SSL: If $\lambda$ is too small, 
the regularization is negligible and the resulting policies do not perform better than the policies generated
by MaxEnt-IRL. On the contrary, if $\lambda$ is too large, the pairwise penalty
forces an extreme smoothing of the reward function and the resulting performance
for both MESSI and MESSIMAX may  decrease.

\textit{Comparison to $\eta$-EM-MaxEnt.}
In our setting, $\eta$-EM-MaxEnt-IRL is worse than MESSI in all but a few iterations, in which it
actually achieves a better performance. The decrease in performance is due to the fact that the unsupervised trajectories are included in
the training set even if they are  unlikely to be generated by the
expert. This is because the estimation of $\theta^*$ is not accurate enough to compute
the actual probability of the trajectories and the error made in estimating these probabilities is amplified over time, leading to significantly worse results.
Finally, one may wonder why
not to interrupt $\eta$-EM-MaxEnt when it reaches its best performance?
Unfortunately, this is not possible since the quality of the policy
corresponding to $\theta_t$ cannot be directly evaluated, which is the same
issue as in SSIRL. On the contrary, since the performance of MESSI is
monotonically increasing, we can safely interrupt it at any iteration.

Our experiments verify the SSL hypothesis (Chapelle et al.~\citeyear{chapelle2006semi-supervised})
that whenever unsupervised trajectories convey some useful information about the
structure of the problem, MESSI performs better than MaxEnt-IRL:
 {\bf 1)} If the expert trajectory is suboptimal and the unsupervised data 
contain the information leading to a better solution. For example, in the
highway experiment, the expert drives only on the left-hand side, while an optimal policy would also use the right-hand side when needed.
{\bf 2)} In the case where the information given by the expert is incomplete and the unsupervised data provide the information about how to act in the
rest of the state space.
On the other hand, if none of the unsupervised trajectories contains useful
information, MESSI performs slightly worse than MaxEnt-IRL. However, our empirical results indicate that the distribution of the trajectories $P_u$ only needs
to partially support $P_{u^*}$ ($\nu \ge 0.15$) for MESSI 
to perform significantly better than MaxEnt-IRL.


\section{Conclusion}


We studied apprenticeship learning with access to unsupervised trajectories in addition to those generated by an
expert. We presented MESSI (MaxEnt Semi-Supervised Inverse reinforcement
learning), that combines MaxEnt-IRL~\cite{ziebart2008maximum}
with SSL by using a pairwise penalty on trajectories. Empirical results showed that MESSI takes
advantage of unsupervised trajectories and can perform better and more
efficiently than MaxEnt-IRL. Moreover, it does not suffer from the disadvantages
of SSIRL (Valko et al.~\citeyear{valko2012semi-supervised}), the only existing semi-supervised
apprenticeship learning algorithm that combines SSL with the IRL approach
of Abbeel and Ng~\shortcite{abbeel2004apprenticeship}.
Possible directions for future research include: {\bf 1)} investigating
other ways to incorporate unsupervised trajectories into the MaxEnt-IRL
framework, {\bf 2)} experimenting with other similarity measures for
trajectories (e.g., the Fisher kernel), and {\bf 3)} studying the
interesting setting in which we have access to more than one expert. 



\newpage

\bibliographystyle{aaai}
\bibliography{Messi}

\newpage

\appendix


\section{Related Work}
\label{sec:related}

\textit{Inverse reinforcement learning} (IRL) as an approach to \textit{apprenticeship learning} was first formally defined in Russell~\shortcite{russell1998learning} as a problem of finding a reward function given the observed trajectories of an agent. In Ng and Russell~\shortcite{ng2000algorithms}, the authors showed that the original problem contains many degenerate solutions such as constant reward functions. They also introduced the assumption of the linearity of the reward function over the state or state-action features, which is still the common assumption in the majority of the IRL approaches.

Many of the IRL methods, such as the one by Abbeel and Ng~\shortcite{abbeel2004apprenticeship},
are based on matching the \textit{expected feature count} of the expert's
trajectories. Their method finds a reward using an iterative
\textit{max-margin} algorithm. This and other max-margin
approaches (e.g.,~Russell~\citeyear{russell1998learning}; Ratliff et al.~\citeyear{ratliff2006maximum}; Syed et al.~\citeyear{syed2008apprenticeship})
aim to find the reward function that makes a deviation
from the expert's policy as costly as possible.

As mentioned in Sec.~\ref{sec:bck}, there are many possibilities of matching
the expert's feature count and the {\em maximum entropy} (MaxEnt) method
of Ziebart et al.~\shortcite{ziebart2008maximum} is one approach to resolve this ambiguity. Several
other methods build on this idea such as those
by Levine et al.~\shortcite{levine2011nonlinear}, Boularias et al.~\shortcite{boularias2011relative}, and Levine and Koltun~\shortcite{levine2012continuous}, as
well as the approach we propose in this paper. In addition to MaxEnt, other
models of the expert behavior have also been investigated in the literature,
such as those by Neu and Szepesv{\'a}ri~\shortcite{neu2007apprenticeship}, Dvijotham and Todorov~\shortcite{dvijotham2010inverse}, and Ramachandran and Amir~\shortcite{ramachandran2007bayesian}.

There have also been other recent approaches to the general apprenticeship learning problem. One approach is to avoid repeatedly solving the forward (direct) RL problem~\cite{boularias2011relative,dvijotham2010inverse,klein2012inverse}. The second one, called \textit{imitation learning}, considers a different setting in which we have access to the expert's policy~\cite{ross2010reduction,bagnell2010efficient,he2012imitation}.

Finally, there are methods that tackle the issue that the expert's
demonstrations could be scarce or costly. These methods include
bootstrapping~\cite{boularias2013apprenticeship}, learning with perturbed
demonstrations~\cite{melo2010analysis}, learning with multiple
rewards~\cite{choi2012nonparametric}, and semi-supervised
learning~\cite{valko2012semi-supervised}. Our algorithm is also based on
semi-supervised learning (SSL), and thus, is related to the work
by Valko, Ghavamzadeh, and Lazaric~\shortcite{valko2012semi-supervised} that combines the IRL approach
of Abbeel and Ng~\shortcite{abbeel2004apprenticeship} with semi-supervised support vector machines
(SVMs). However, unlike their work, ours does not heavily rely on the {\em
cluster assumption}, i.e., a common distributional assumption in SSL, which
assumes that the good and bad trajectories should be well-separated in some
feature space defined over the state space. The biggest disadvantage of their
method is that it eventually converges to the supervised case and it is unknown
whether a good stopping 
criterion could be designed. Another problem is that their objective function is non-convex and very difficult to optimize. 


\section{Additional Experiments}\label{s:add.experiments}

In this section, we report additional empirical results which mostly confirm the findings reported in the main paper.

 \subsection{The Grid-world Problem}\label{sub:grid}

\begin{figure}[ht]
     \begin{center}
 \includegraphics[trim = 10mm 1mm 5mm 1mm,scale=0.21]{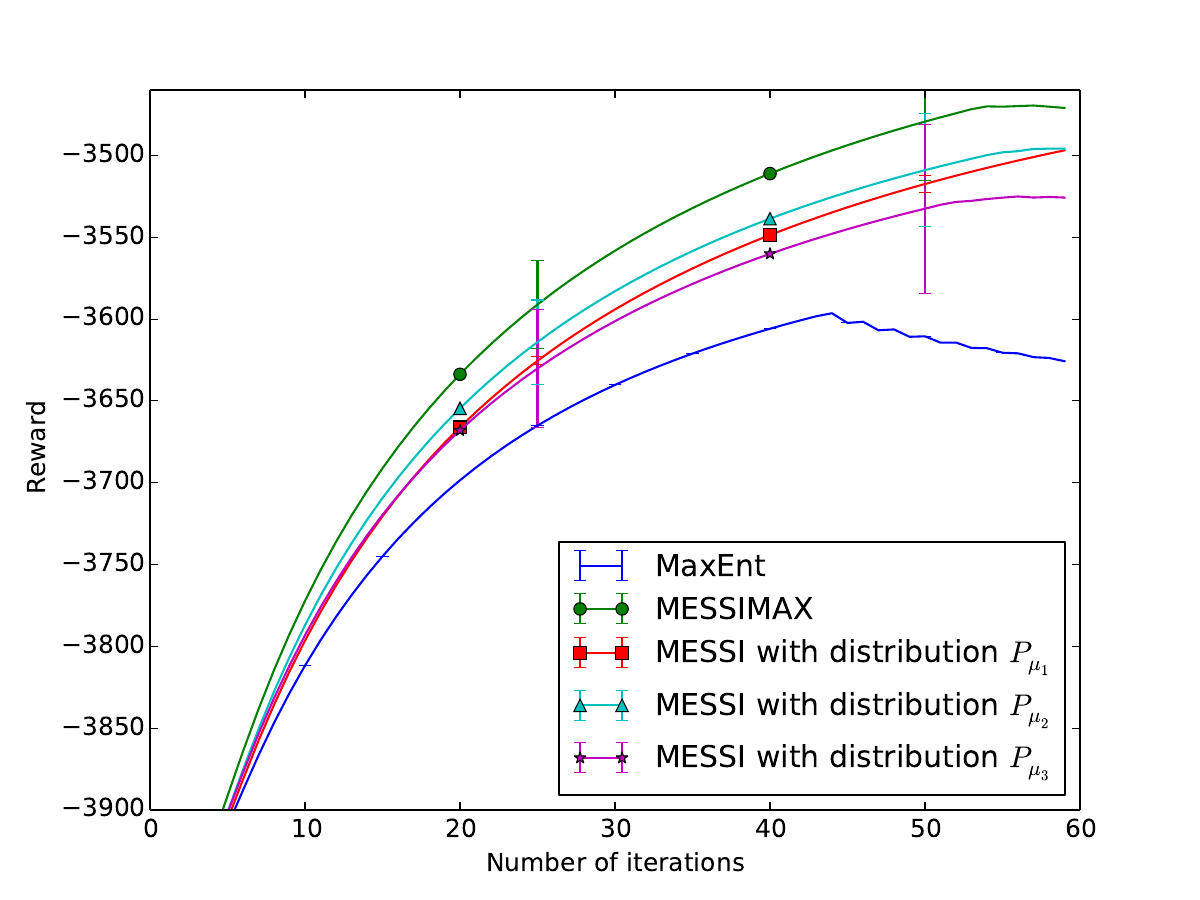}
 \includegraphics[trim = 1mm 1mm 10mm 1mm,scale=0.21]{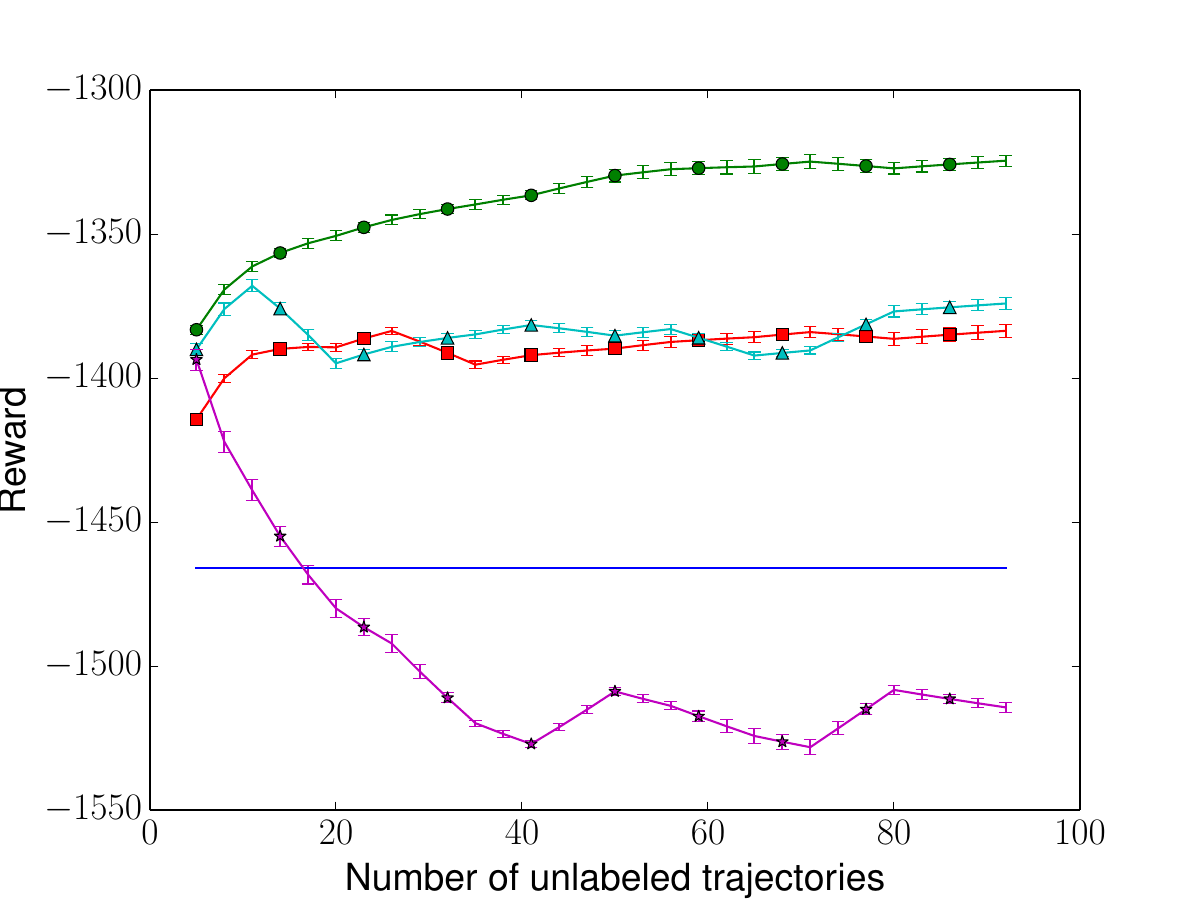}
 \includegraphics[trim = 10mm 1mm 5mm 1mm,scale=0.21]{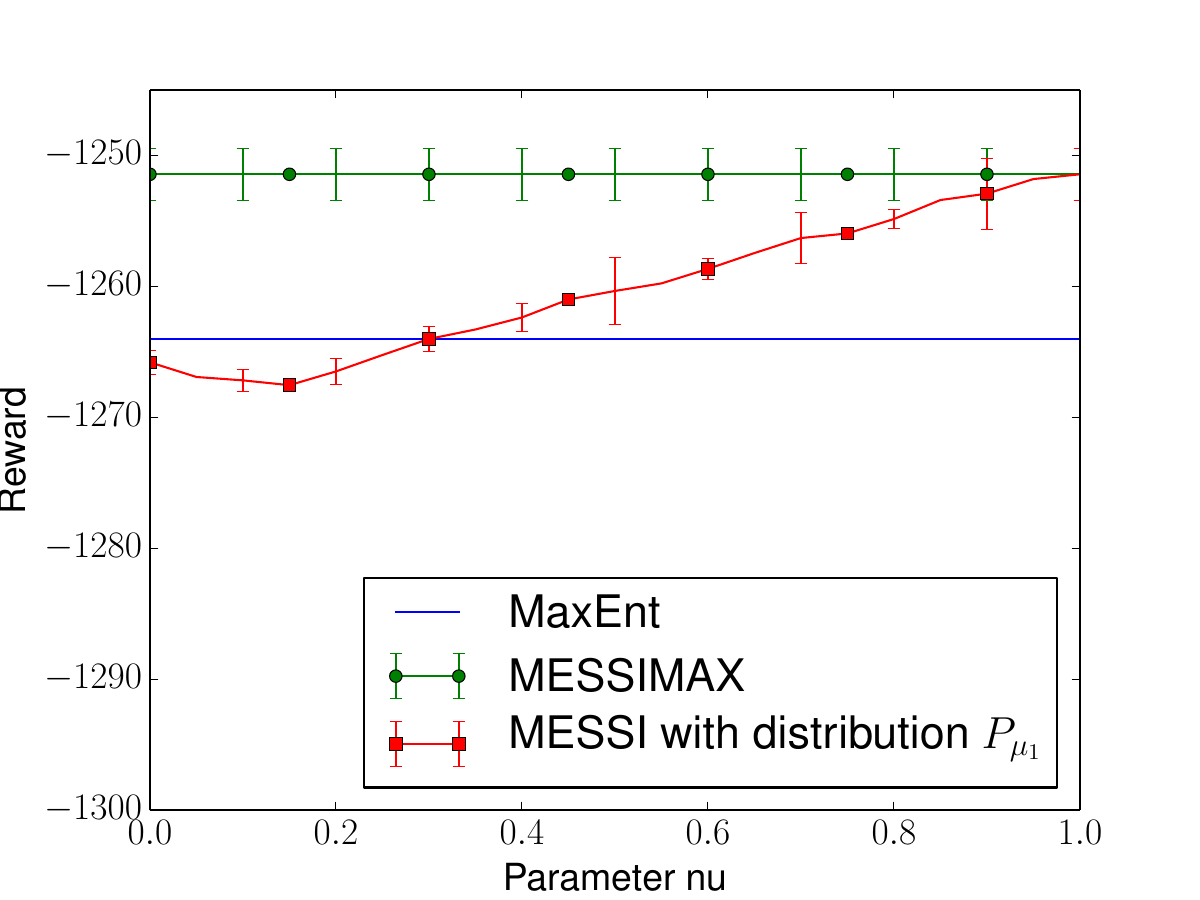}
 \includegraphics[trim = 1mm 1mm 25mm 1mm,scale=0.21]{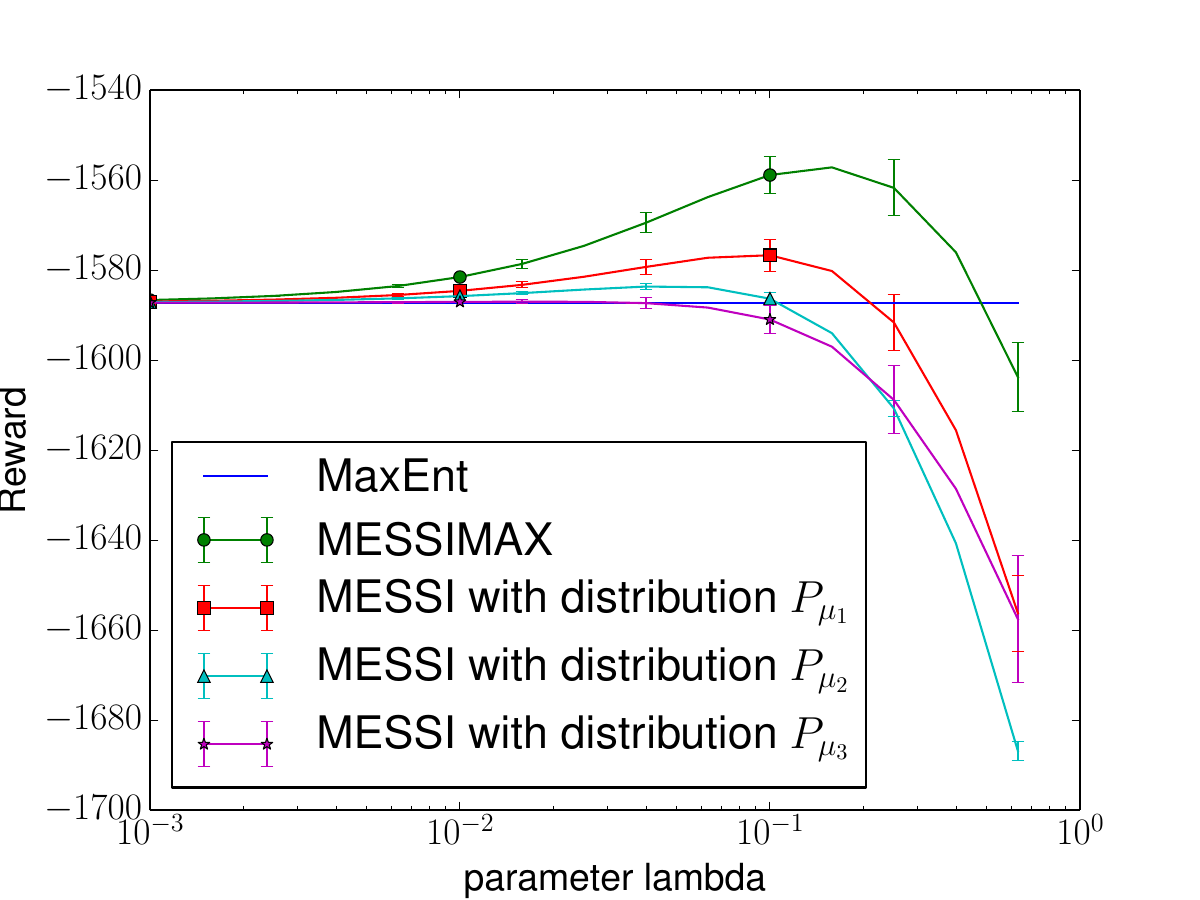}
 \caption{Results on the grid-world problem as a function of (from left to right): number of iterations of the algorithms, number of unsupervised trajectories, parameter $\nu$, and parameter $\lambda$.}
\label{fig:xpgrid}
  \end{center}
 \end{figure} 
 
  \begin{figure}[ht]
     \begin{center}
 \includegraphics[trim = 25mm 1mm 10mm 1mm,scale=0.15]{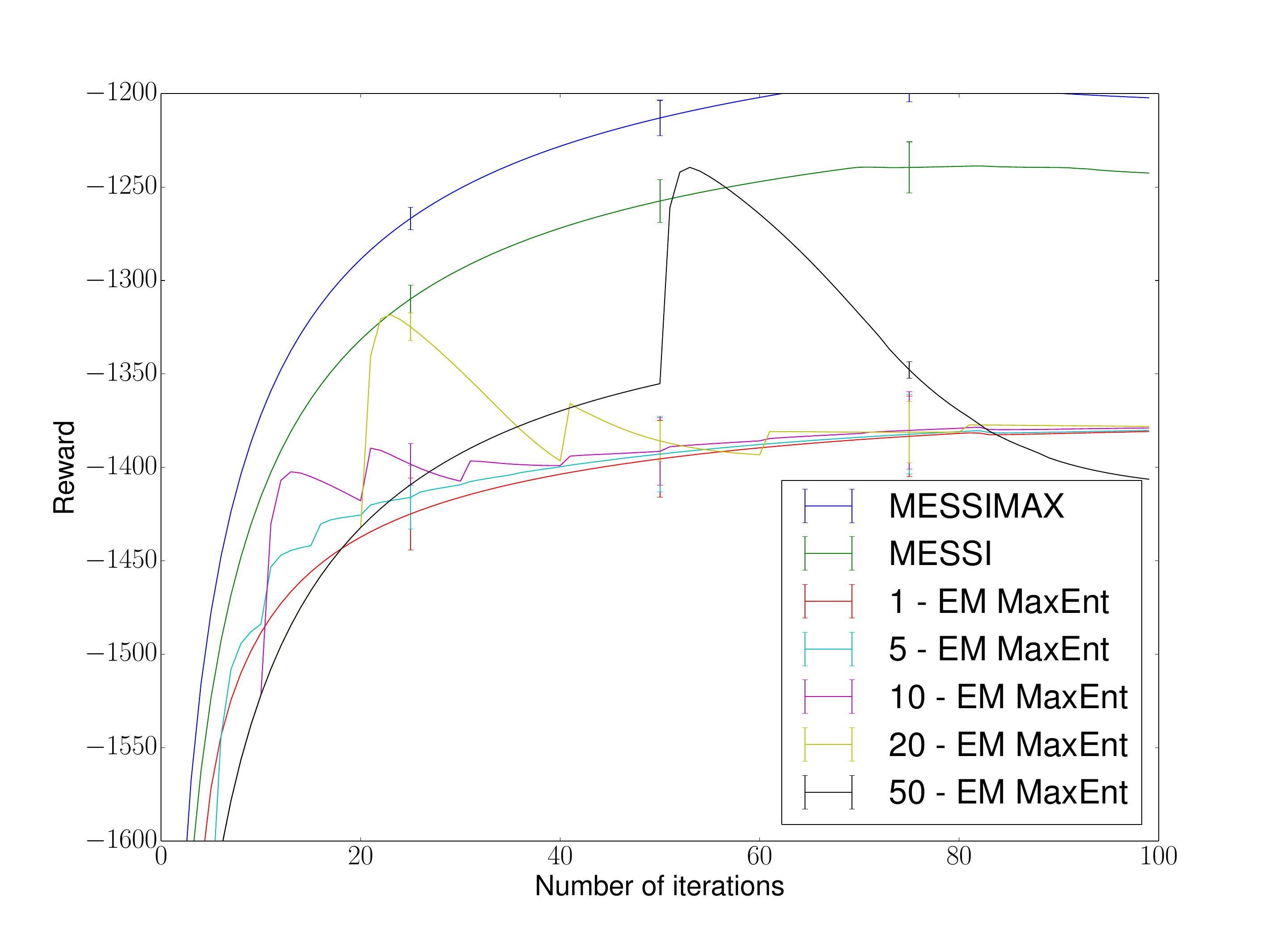}
\caption{Comparison between the performance of MESSI and MESSIMAX with the iterative $\eta$-EM-MaxEnt with $\eta=1, 5, 10, 20, 50$ on the grid-world problem.}
\label{fig:comparison.em.grid}
  \end{center}
 \end{figure}
 
In this set of experiments, we use a $16\times 16$ square grid-world, a variation of the grid-world domain in Abbeel and Ng~\shortcite{abbeel2004apprenticeship}. The agent has $4$ possible actions (\textit{up}, \textit{down}, \textit{left}, \textit{right}) with $70$\% chance of success and $30$\% chance of taking a random different action. The grid is divided into $64$ macro-states of size $2\times 2$ each, which define the feature space $\boldf$, so that each macro-state has its own characteristic feature that is active when a state within a macro-state is traversed.
A reward vector $\btheta^*$ is generated at random and is set as the
\textit{true} reward of the problem. This reward maps every feature to a
strictly negative value, except for $3$ randomly chosen features. 
$\btheta_1$  and $\btheta_2$ are also generated at random. 
%
In this problem, we use the RBF kernel $s(\zeta_i,\zeta_j)=\exp(-\| \boldf_{\zeta_i} - \boldf_{\zeta_j} \|/10)$ to define the similarity between two trajectories $\zeta_i$ and $\zeta_j$. This similarity function is completely general and does not exploit any prior knowledge about the problem. We evaluate the performance of a policy as its expected reward w.r.t.~the true reward, i.e.,~$\btheta_{\text{true}}^{\top} \boldf_T$ .
Results are reported in Fig.~\ref{fig:xpgrid}. The result of the comparison with the $\eta$-EM-MaxEnt are reported in Fig.~\ref{fig:comparison.em.grid}.

\subsection{The Pit Problem}\label{sub:pit}

   \begin{figure}[th]
     \begin{center}
 \includegraphics[width=0.3\columnwidth]{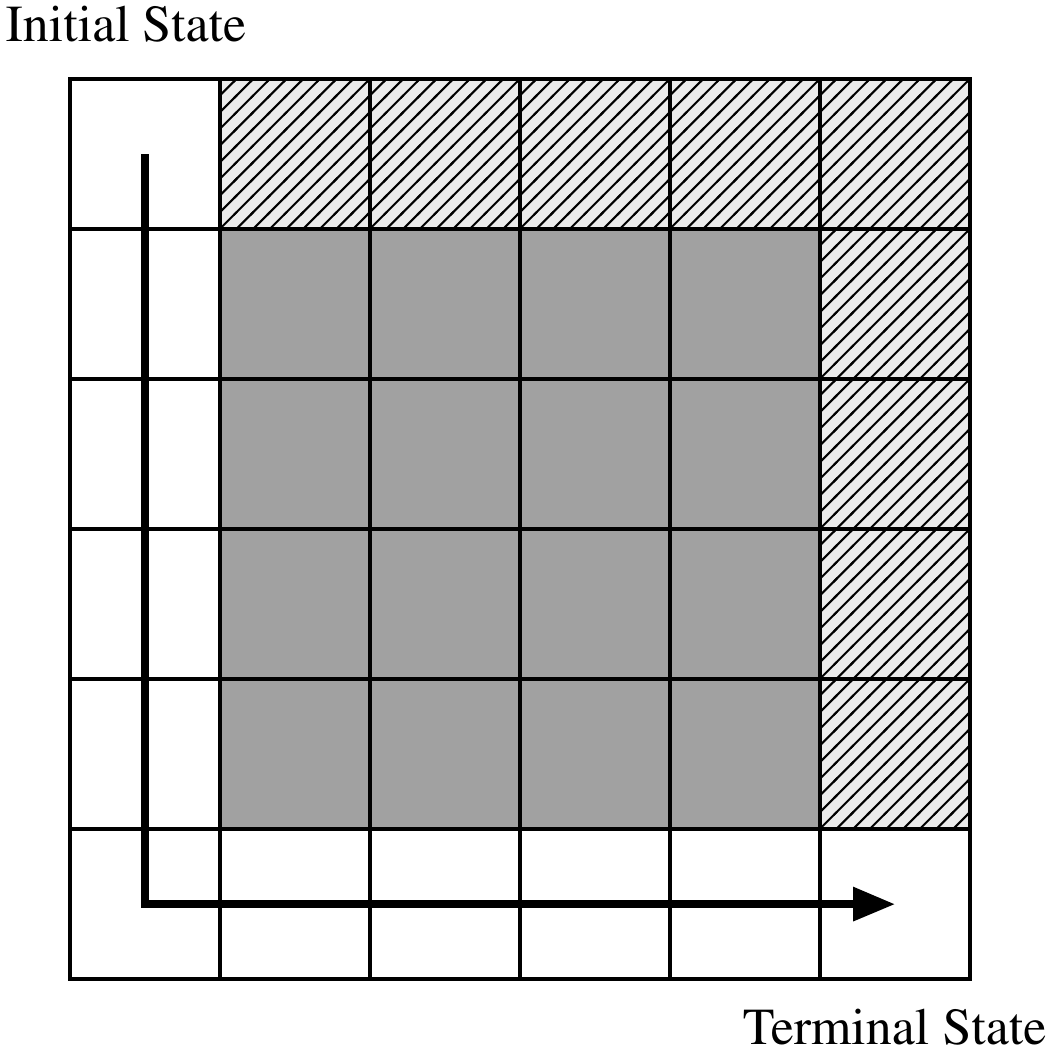}
 \caption{ The pit domain.}
 \label{fig:pit}
  \end{center}
 \end{figure}

\begin{figure}[ht]
     \begin{center}
 \includegraphics[trim = 30mm 1mm 20mm 1mm,scale=0.21]{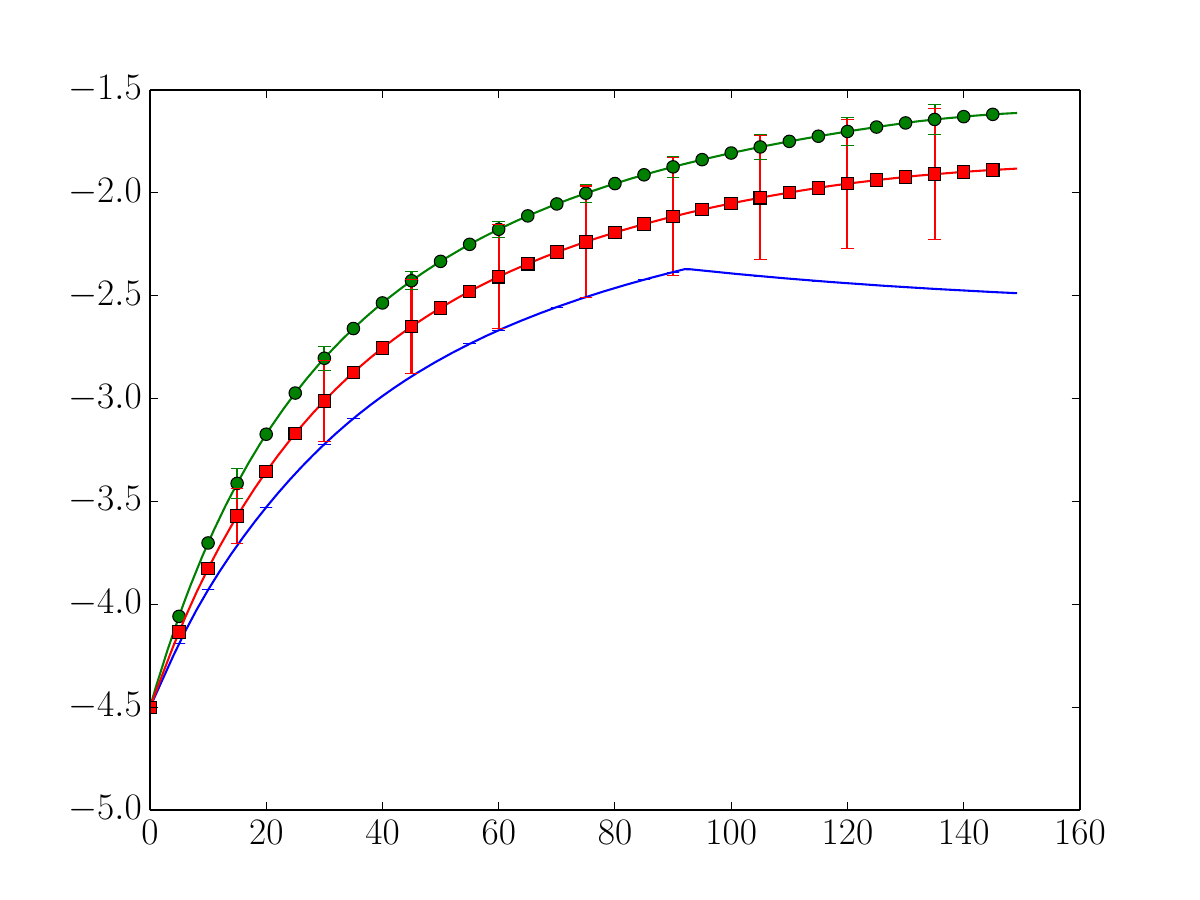}
  \includegraphics[trim = 1mm 1mm 10mm 1mm,scale=0.21]{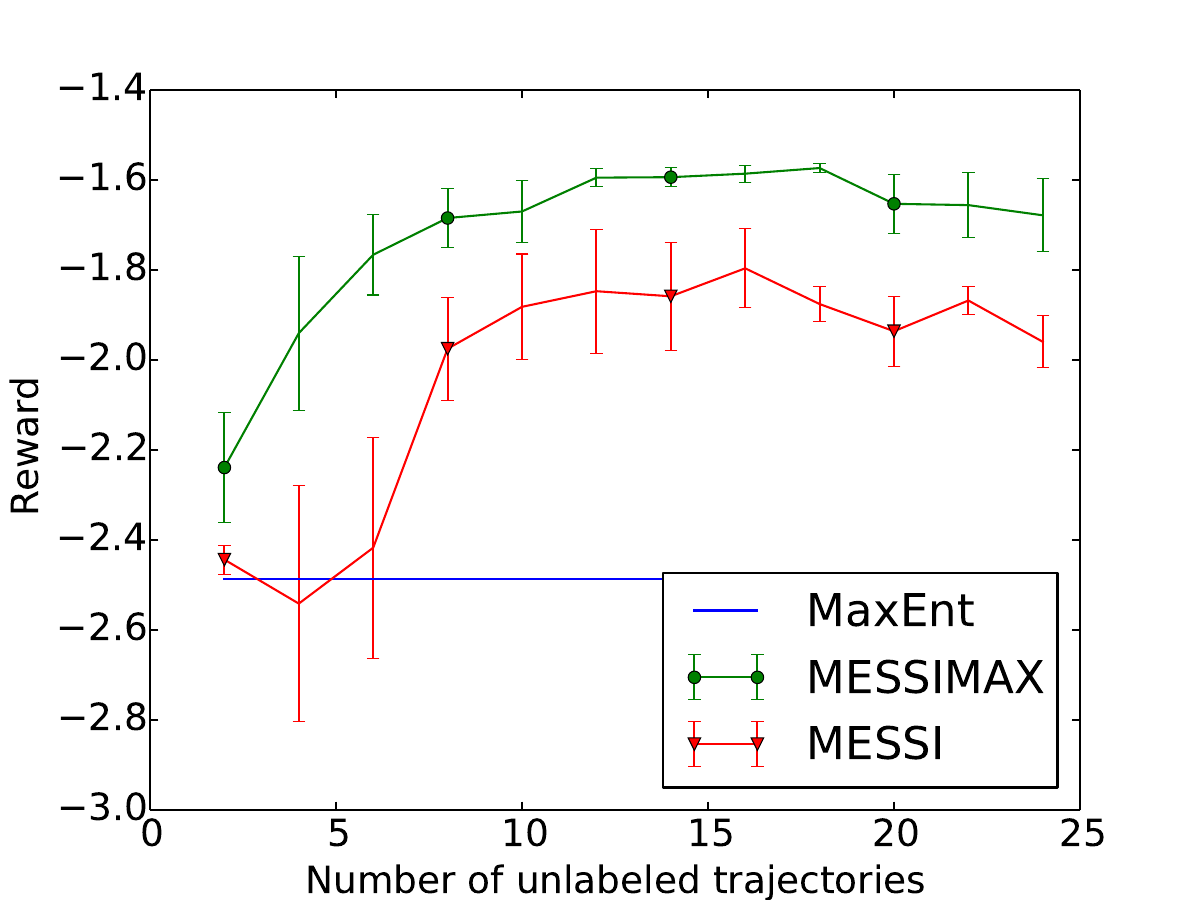}
 \includegraphics[trim = 10mm 1mm 10mm 1mm,scale=0.21]{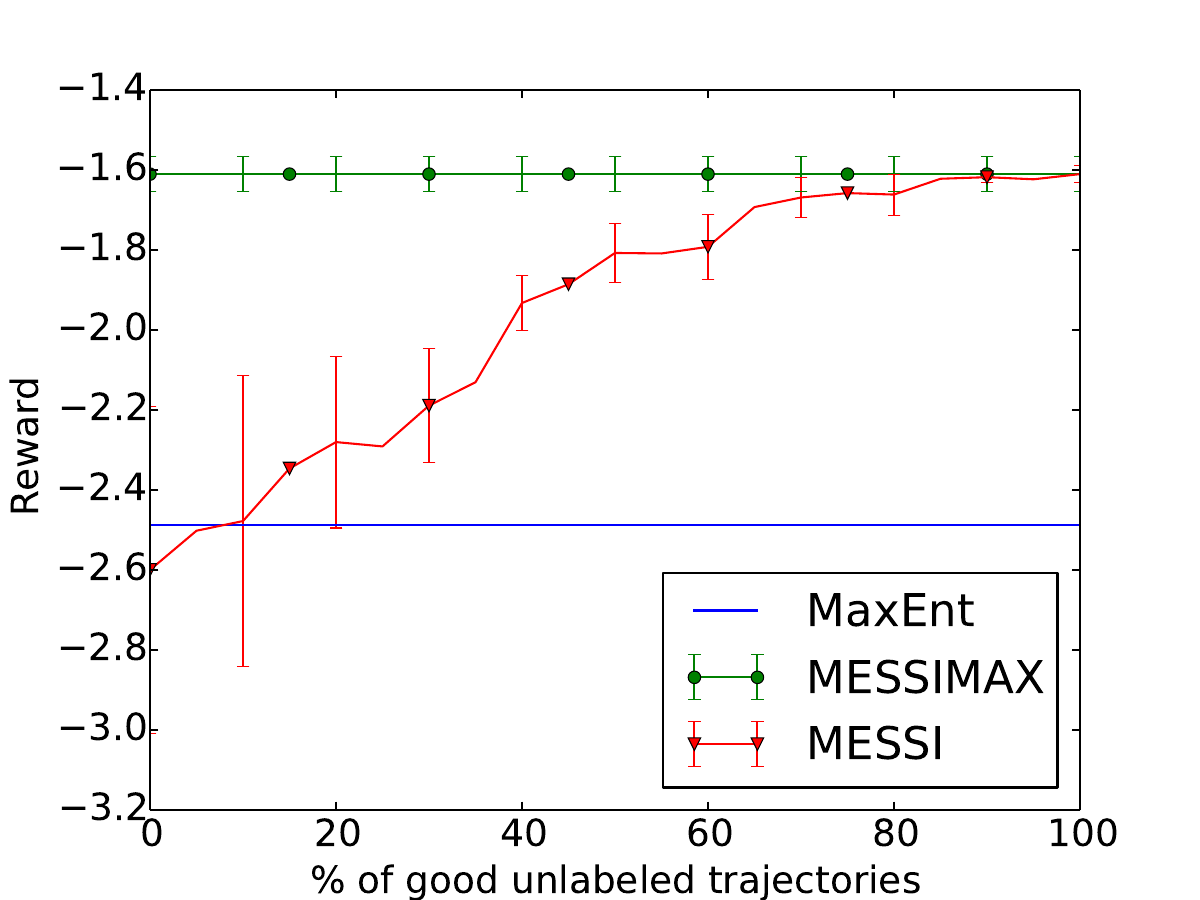}
 \includegraphics[trim = 1mm 1mm 25mm 1mm,scale=0.21]{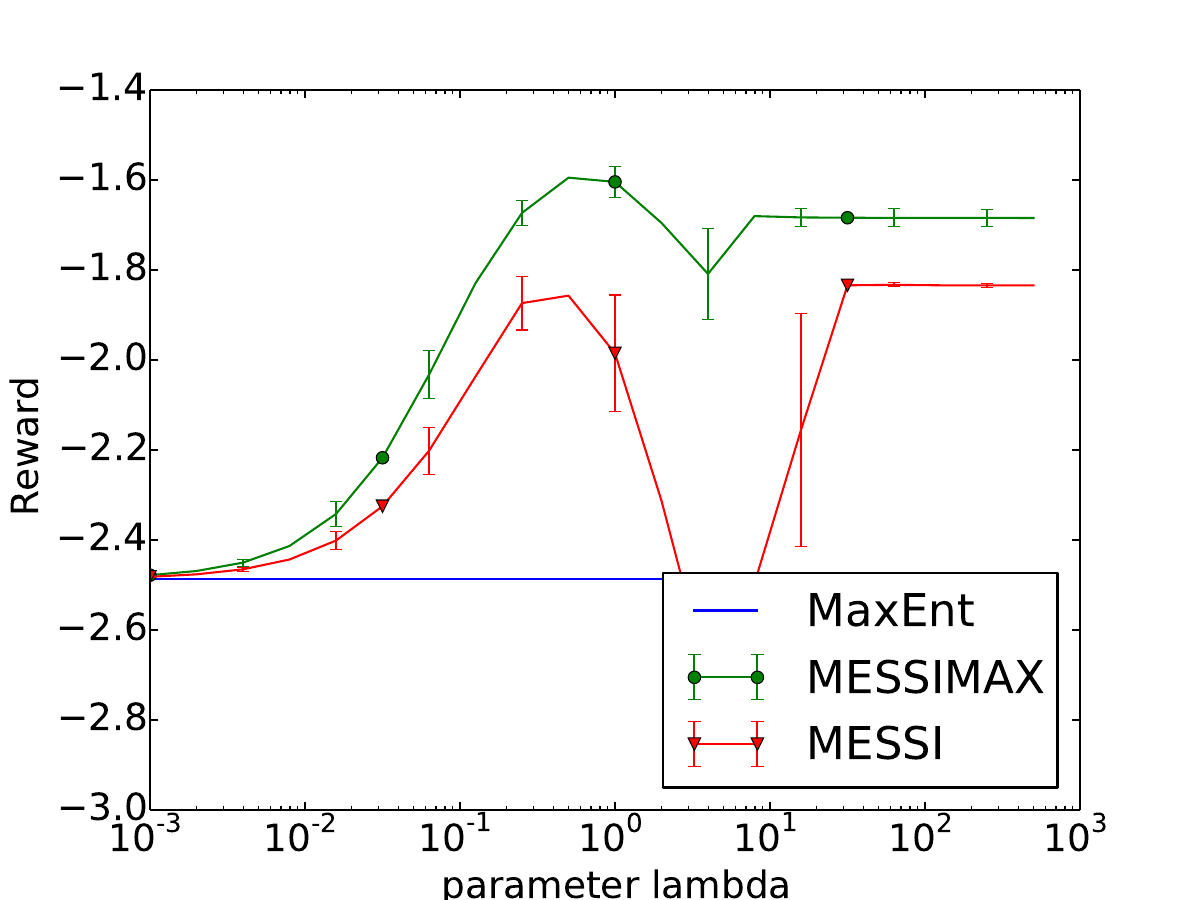}
 \caption{Results on the pit problem as a function of (from left to right): number of iterations of the algorithms, number of unsupervised trajectories, parameter $\nu$, and parameter $\lambda$.}
\label{fig:xppit}
  \end{center}
 \end{figure}
 
In this set of experiments, we use a $6 \times 6$ grid world representing an edge surrounding a pit (see Fig.~\ref{fig:pit}). The agent has $4$ possible actions (\textit{up}, \textit{down}, \textit{left}, \textit{right}) with $85$\% chance of success and $15$\% chance of taking a random different action. The initial state is the square with coordinate $(1,1)$ and the terminal state  is the square with coordinate $(6,6)$. The grid is divided in $3$ parts that correspond to the $3$ different features, the left edge (represented by the white squares in Fig.~\ref{fig:pit}), the right edge (shaded squares), and the pit (the gray squares in the middle). The objective of the expert is to move from the initial square to the terminal state by avoiding the pit in the middle. The (single) expert trajectory (the black arrow) provided to the learner goes around the pit counter-clockwise. In this case, the unsupervised trajectories are generated as a mixture of trajectories generated from a deterministic policy that goes around the pit either clockwise or counter-clockwise (basically $P_{u^*}$) and trajectories are generated from a policy that crosses the pit.
%
%
The similarity function used in this experiment is hand-crafted for this domain and is defined as $s(\zeta_j,\zeta_k)=\exp(-\| n_j - n_k \|)$, where $n_j$ denotes the number of change of direction in the trajectory $\zeta_j$ (for instance going left then down). This is an example of a hand-crafted similarity function that fits nicely to the problem, since a trajectory is good if and only if it goes around the pit, and thus, turns only once.
Since a good trajectory should carefully avoid the pit, for this experiment, we evaluate the performance of a policy by how frequently it crosses the pit, that corresponds to evaluating the expected feature count $\boldf_T$ (obtained after $T$ iterations of gradient descent) corresponding to the pit. We denote by $\boldf_T^{\text{pit}}$ this value, and report $-\boldf_T^{\text{pit}}$ as a measure of performance in the plots of Fig.~\ref{fig:xppit}. The results basically confirm the discussion in Sec.~\ref{s:experiments} and show an even stronger advantage of MESSI w.r.t.~MaxEnt-IRL, i.e.,~whenever prior knowledge about the problem is available and a very informative similarity function is chosen, MESSI is very effective in taking advantage of it and significantly outperforms MaxEnt-IRL.

\end{document}